\documentclass[fleqn,10pt]{wlscirep}
\usepackage[utf8]{inputenc}
\usepackage{newtxmath}
\usepackage[T1]{fontenc}

\title{DiffDTM: A conditional structure-free framework for bioactive molecules generation targeted for dual proteins.}

\author[1,2]{Lei Huang}
\author[2]{Zheng Yuan}
\author[3]{Huihui Yan}
\author[3]{Rong Sheng}
\author[1]{Linjing Liu}
\author[1]{Fuzhou Wang}
\author[1]{Weidun Xie}
\author[1]{Nanjun Chen}
\author[2]{Fei Huang}
\author[2]{Songfang Huang}
\author[1,*]{Ka-Chun Wong}
\author[2,*]{Yaoyun Zhang}
\affil[1]{City University of Hong Kong, Hong Kong SAR}
\affil[2]{Alibaba Damo Academy, China}
\affil[3]{Zhejiang University, China}

\affil[*]{zhangyaoyun.zyy@alibaba-inc.com}

\begin{abstract}
Advances in deep generative models shed light on \textit{de novo} molecule generation with desired properties. However, molecule generation targeted for dual protein targets still face formidable challenges including protein 3D structure data requisition for model training, auto-regressive sampling, and model generalization for unseen targets. Here, we proposed DiffDTM, a novel conditional structure-free deep generative model based on a diffusion model for dual targets based molecule generation to address the above issues. Specifically, DiffDTM receives protein sequences and molecular graphs as inputs instead of protein and molecular conformations and incorporates an information fusion module to achieve conditional generation in a one-shot manner. We have conducted comprehensive multi-view experiments to demonstrate that DiffDTM can generate drug-like, synthesis-accessible, novel, and high-binding affinity molecules targeting specific dual proteins, outperforming the state-of-the-art (SOTA) models in terms of multiple evaluation metrics. Furthermore, we utilized DiffDTM to generate molecules towards dopamine receptor D2 and 5-hydroxytryptamine receptor 1A as new antipsychotics. The experimental results indicate that DiffDTM can be easily plugged into unseen dual targets to generate bioactive molecules, addressing the issues of requiring insufficient active molecule data for training as well as the need to retrain when encountering new targets.    
\end{abstract}
\begin{document}

\flushbottom
\maketitle
%
%
\thispagestyle{empty}


\section*{Introduction}

Drug molecule generation targeting for specific protein pockets is an essential task in drug discovery. In the past, traditional drug discovery attempts to develop selective drug candidates with high affinities with a single specific target protein. The “one drug, one target” paradigm could provide highly potent and specific treatment since they have no off-target effects\cite{talevi2015multi}. However, modulating one target is hard to heal and alleviate the complex disorders in which the breakdown of robust physiological systems results from a combination of multiple genetic and/or environmental factors \cite{ay2007drug}. Besides, drug molecules targeting at a single target are prone to drug resistance or side effects because a single target may be involved in multiple physiological functions. Instead, most therapeutics produce their desired effects by modulating multiple targets and pathways \cite{ramsay2018perspective}\cite{saginc2017harnessing}. Therefore, drug discovery for multiple targets which is called polypharmacology has emerged as a promising paradigm for treating complex disorders by regulating multiple targets to accomplish the intended molecular reactions \cite{anighoro2014polypharmacology}. Multi-target based ligands avoid the risk of drug-drug interactions compared to other multi-target therapies, such as the drug cocktail (refers to the simultaneous use of multiple drugs) and multi-component drug (refers to the drug composed of multiple active ingredients) \cite{edwards2000adverse}. Thus, there is a growing need to find versatile candidate molecules which could simultaneously regulate multiple targets of the same complex disease.

However, naive exhaustive searches in the massive chemical space (range from $10^{60}$ to $100^{100}$ depending on the size of desired molecules)\cite{Satyanarayanajois2011} are prohibitively costly. In addition, the traditional workflow is bottlenecked by historical knowledge, thus infeasible to explore novel molecular structures which are not recorded in the existing databases\cite{Huang2023.01.28.526011}. Fortunately, the development of computer-aided drug design (CADD) has paved the way for efficient novel drug design by learning the inherent density of real-world molecule data with prior knowledge. This method plays a portal role in the drug development process, which can reduce experiments and costs and improve the efficiency and success rate of drug development. In the meantime, the growing data amount sheds light on the application of machine learning methods in drug discovery. The exiting machine learning methods have been developed for prediction drug property\cite{liu2019chemi}\cite{zhang2021motif}, drug-drug interaction\cite{huang2022egfi}\cite{lin2020kgnn}\cite{pang2022amde}, and drug-target interaction\cite{huang2022coadti}\cite{moon2022pignet}, archiving optimal performances. 

Specifically, deep generative models have been blossoming for novel drug generation with desired properties in recent years. The current drug generation methods utilize the simplified molecular input line entry system (SMILES) representations or molecular graphs to encode the drug molecule as the inputs of models\cite{gomez2018automatic} \cite{zang2020moflow} \cite{guo2021data} \cite{grisoni2020bidirectional}  \cite{luo2021graphdf} \cite{shi2020graphaf}. By employing sequential models such as recurrent neural network (RNN) and transformer, those methods are capable of learning the sequential information from SMILES sequences. On the other hand, the models design the generative framework based on graph neural network (GNN) to learn the graph-structured data. Although SMILES representation is able to imply chemical structures implicitly, molecule graphs whose atoms and chemical bonds are represented by nodes and edges can indicate the molecule structures explicitly. Building on the variational autoencoder (VAE)-based models\cite{jin2018junction}, generative adversarial network (GAN)\cite{de2018molgan}, normalizing flows\cite{zang2020moflow}\cite{luo2021graphdf} and diffusion models\cite{huang2022mdm}, the generative models which focus on generating molecular graphs generate the atom types and the corresponding chemical bonds in one-shot or auto-regressive manners. There are also researches incorporating 3D information to generate molecules geometries which represent the molecules as point clouds and construct the chemical bonds by atomic pairwise distances\cite{huang2022mdm}\cite{gebauer2019symmetry}\cite{garcia2021n}\cite{hoogeboom2022equivariant}\cite{yuan2023molecular}. Followed structure-based methods\cite{Huang2023.01.28.526011}\cite{luo20213d}, then incorporate the protein structure information as context information to enable the model to learn the pocket cavity, making the molecules adapted for the required structure. All of those methods are claimed to generate novel drug-like molecules efficiently and accurately. Despite the advantages, those methods suffer three limitations. First, they are not designed to generate molecules targeting specific proteins or only targeting a single protein. Instead, those methods generate the molecules from scratch, which sample from the predefined noise distribution. Thus these generated molecules cannot manipulate the specific biology functions and ensure the therapeutic effect. Second, most of them generate molecules autoregressively, suffering from deviation accumulations especially when invalid structures are generated in the early step. Finally, the structure-based methods require a large amount of three-dimensional protein and molecule structures. However, co-crystal structures of protein–ligand pairs are hard to be verified by researchers while high dimensional structure matrices are always sparse, leading to tedious and redundancy computations \cite{zheng2020predicting}.

The recently developed methods which could generate molecules adapted for dual proteins utilize molecular SMILES as inputs and then train the models on the data which are active towards both targets\cite{lu2021novo}. However, only utilizing SMILES sequences cannot model the molecular structure explicitly as we mentioned. In addition, they require a large amount of molecule data which are bioactive towards the dual targets to train the model and do not incorporate protein information directly. Thus, their models are limited by the data of specific dual targets. Their models lack generalizability as they require retraining for every novel pair of targets. In this regard, we proposed a novel conditional structure-free deep generative model based on a Diffusion model for Dual Targets based Molecule generation (DiffDTM) to tackle the above issues. DiffDTM requires molecule graphs and protein amino acid sequences rather than the 3D structures as inputs, which could leverage large amounts of amino acid sequence data without the limitations of tertiary structure data and predict the function and structure of molecules without explicitly considering the constraints of three-dimensional space. Based on the diffusion model tailored for modelling molecule data distribution, DiffDTM could generate molecules in one shot manner, enabling it to consider global information and avoid deviation accumulation. Then, we built a graph neural network to represent molecules as it can incorporate the structure information by introducing inductive bias, enhancing the ability to learn the functional relevance of latent space. By incorporating the pre-trained model esm2\cite{lin2023evolutionary} to encode protein amino acid sequences, DiffDTM is capable of fusing the protein information into the process of molecule generation. DiffDTM could generate corresponding bioactive molecules for any given pair of protein targets, which means the model does not require retraining for every protein pairs. To the best of our knowledge, we are the first to design a deep generative model for molecule generation given arbitrary dual targets. Working on extensive collection integrated from multiple data sources, we demonstrate that DiffDTM outperforms state-of-the-art (SOTA) models in terms of binding affinity scores and achieves comparable performance on multiple metrics.

\section*{Results}
\subsection*{The overall architecture of conditional structure-free deep generative framework for bioactive molecules generation targeted for dual proteins}
\begin{figure}
    \centering
    \includegraphics{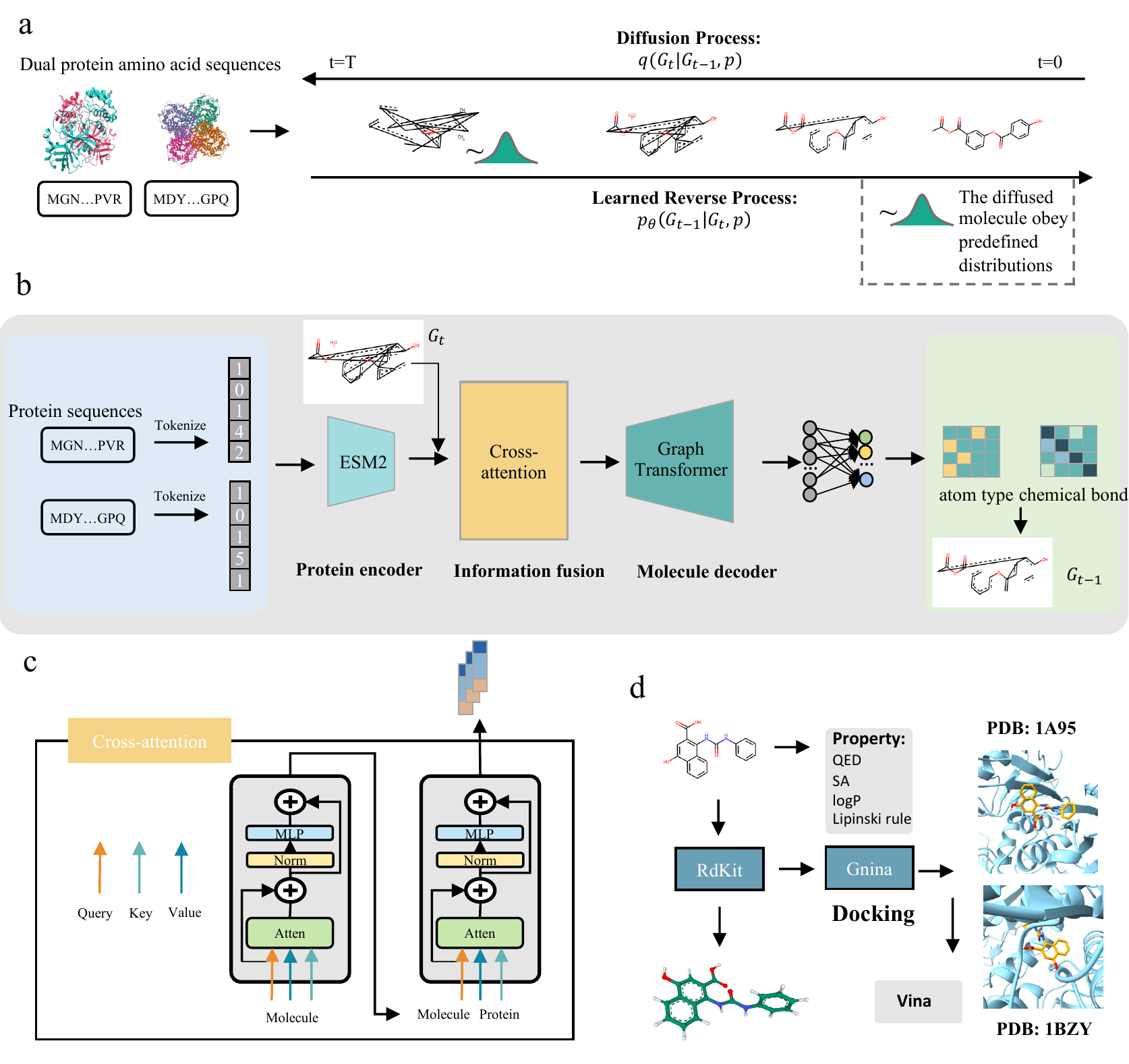}
    \caption{Overview of the DiffDTM model. (a). The diagram of the diffusion model of DiffDTM. DiffDTM defines the diffusion process and the reverse process. The arrow pointing from right to left represents the diffusion process, and the arrow pointing from left to right represents the reversed process. The diffusion process gradually adds noise to the molecule data while the learned reverse process utilizes a neural network $\phi_\theta$ to predict the denoised molecule data to reverse the diffusion process given conditioned protein information. (b). The model structure of DiffDTM. DiffDTM is a structure-free model which only receives protein sequences as inputs. Then the protein sequences are tokenized to feed into a pre-trained protein language model ESM2. DiffDTM inputs the noisy molecule data $G_{t-1}$ in the last time step and then fuses the molecule data with protein embeddings in cross-attention layers. The enriched molecule representations are fed into graph transformer and subsequent MLPs to predict the denoised atom types and chemical bonds to obtain the data point $G_{t}$ in current time step. (c). The illustration of cross-attention layers for information fusion. Cross-attention layers consist of two transformer layers. The first transformer implements self-attention for the molecule data. The second transformer is designed to measure the effects of proteins on molecules. d. The process of molecule bioactive evaluation. Firstly we use RdKit to predict the molecular confirmations. We then employ Gnina to predict the docking pose to calculate vina scores. We also use RdKit to calculate the chemical property of generated molecules, including QED, SA, logP and Lipinski rules.}
    \label{fig:model}
\end{figure}

The proposed framework DiffDTM is illustrated in Figure \ref{fig:model}. DiffDTM is based on the diffusion model\cite{sohl2015deep}. The diffusion model defines two Markov processes: diffusion process and reverse process. As presented in Figure \ref{fig:model}a, the diffusion process gradually adds Gaussian noise to diffuse the real molecule data distribution into a predefined noise distribution while the reverse distribution aims to utilize machine learning models to reverse the diffusion process to recover the noise sampled from the predefined noise distribution to real molecule data given dual protein information. Specifically, we present the machine learning model incorporated in the diffusion framework in Figure\ref{fig:model}b. First, we utilize a pre-trained model ESM2 which is a transformer protein language model pre-trained on Unified Resolution 50 dataset to encode the protein sequences. In this way, DiffDTM could harness external knowledge to enrich the protein representation. Given the diffused molecule data, we then designed cross-attention layers to fuse the protein information and molecule information. 

The cross-attention layer consists of two transformer layers. In the first transformer layer which can be considered as a feature encoder, only molecules are inputted into it to obtain the molecule feature representation. After exploiting the molecule feature representation and protein representation, DiffDTM translates the protein modality to molecule modality in the second transformer layer, which can measure the effects of proteins on the molecule. In other words, this design partially simulates the drug-target interaction by utilizing the attention score in the latent space. We named DIffDTM with this strategy as DiffDTM-CA. In addition, we designed two other different methods of information fusion. We select the CLS feature vector outputted from EMS2 model, which is inferred by other amino acid words and thus could represent the sequence information. Then we concatenate the CLS vector with molecule vectors in the feature dimension. We named DIffDTM with this strategy as DiffDTM-CAT. For the last strategy, we treat the protein as a virtual node and construct virtual edges between the virtual node and all the atoms of the molecule. We named DIffDTM with this strategy as DiffDTM-VN.

The molecule representations with protein information outputted from the cross-attention layers are then fed into the graph transformer, which could learn the molecule graph feature through the message passing mechanism. Then We could obtain the atom type and chemical bond noises from the following MLPs. Ultimately, the removal of noise from the molecular data at the present time step allows for the acquisition of the molecular data at the subsequent time step.

\subsection*{DiffDTM can generate valid, novel, drug-like bioactive molecules towards dual tagrtes}
We extensively assessed the generation performance of DiffDTM on multiple metrics. Since we are the first to generate molecules given arbitrary dual targets, there are no baseline models for comparison. Therefore, we design three baseline models based on GraphVAE, JTVAE\cite{jin2018junction} and MolGPT\cite{bagal2021liggpt} for performance comparison. Here, we adopt nine metrics to extensively evaluate the qualities of generated molecules: Validity, Uniqueness, Novelty, SA (synthetic accessibility), QED, logP, Lipinski rule, Toxicity score and Vina score.  In order to avoid data leakage, We selected 100 dual targets that do not appear in the training set and whose sequence identity is less than 10\%. Finally, we generate 100 molecules for each dual target (10000 molecules in total).

\begin{figure}
    \centering
    \includegraphics{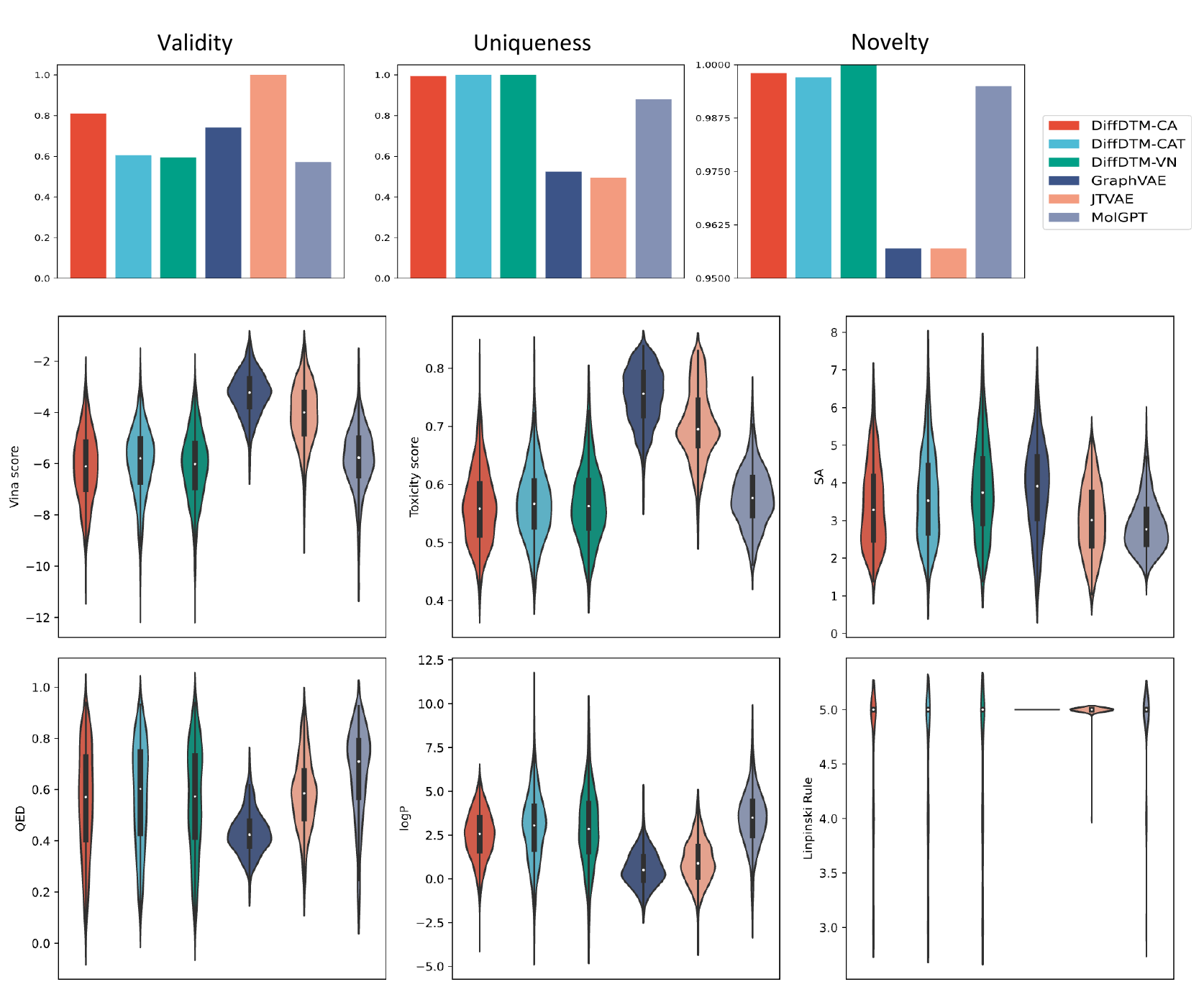}
    \caption{Performance of DiffDTM and baseline models on the independent test set. We report the results of DiffDTM-CA, DoffDTM-CAT, DiffDTM-VN, GraphVAE, JTVAE, and MolGPT on validity, uniqueness, novelty, vina score, toxicity score, sa, QED, logP, and Lipinski rules.}
    \label{fig:performance}
\end{figure}

Figure \ref{fig:performance} presents the overall performances of generated molecules of DiffDTM and baseline models. For the quantitative metrics, DiffDTM-CA achieves the second highest validity (0.809) except for JTVAE, which constructs defined fragments to assemble molecules. On the other hand, MolGPT has the lowest validity score. Despite the capability of JT-VAE to generate valid molecules, it is limited in generating repeated molecules. Similarly, GraphVAE also has limited power to generate unique molecules. Instead, DiffDTM-CA and its variants could generate unique molecules, which reach 0.995,1,1 and exceeds baseline models by at least 13.2\% on uniqueness. Furthermore, DIffDTM can generate more novel molecules which do not exist in the training dataset than the baseline models, reaching 0.998,0.997,1 on novelty.

Since we focus on generating molecules targeting dual targets, it is essential to evaluate the binding affinities of the molecules and dual targets. Here, we report the average of the vina scores of molecules with the two targets. As we can observe in Figure \ref{fig:performance}, DiffDTM-CA and its variants achieve the lowest vina score (-6.123,-5.932, and 6.092) compared to the baseline models, indicating that DiffDTM could generate molecules with high affinities with the targeted proteins. In drug development, besides considering factors such as drug-target affinity and efficacy, drug safety is also an important aspect. Among them, drug toxicity is one of the key indicators for evaluating drug safety. We notice that the molecules generated by DiffDTM-CA have the lowest toxicity score(0.561), which means DiffDTM has learned the rational structure of the molecules and generates less toxic structures. In addition, we also report the physicochemical property distributions of the generated molecules. As we can observe, the SA score of DiffDTM generated molecules is comparable to the baseline models. Besides, DiffDTM could generate molecules with higher QED, and appropriate logP values, and most of the molecules obey the Lipinski rules. In general, DiffDTM could generate valid, novel, drug-like bioactive molecules towards dual targets by adopting the information fusion models to incorporate protein information and eliminating the learned noise based on the diffusion model during sampling.

\subsection*{DiffDTM enables bioactive molecules generation towards DRD2 and HTR1A}
Since DiffDTM can be easily extended to an arbitrary pair of proteins and does not require bioactive data for training, we study Dopamine receptor D2 (DRD2) and 5-hydroxytryptamine receptor 1A (HTR1A) as the dual targets which are not in training set to investigate whether DIffDTM could generate bioactive molecules for unseen targets. DRD2 and HTR1A are neurotransmitter receptors which are commonly studied as dual targets in drug development. DRD2 is a D2-type dopamine receptor that belongs to the class of G protein-coupled receptors and is widely distributed in the central nervous system. HTR1A is a 1A-type serotonin receptor that belongs to the G protein-coupled receptor family and is mainly distributed in regions such as the amygdala, hippocampus, and prefrontal cortex. Dual-target therapy involving DRD2 and HTR1A means that a single drug can simultaneously regulate the activity of both receptors. This dual-target has been shown to have potential therapeutic effects on various diseases, such as schizophrenia, emotional disorders, and drug addiction. 

\begin{figure}[htbp]
    \centering
    \includegraphics{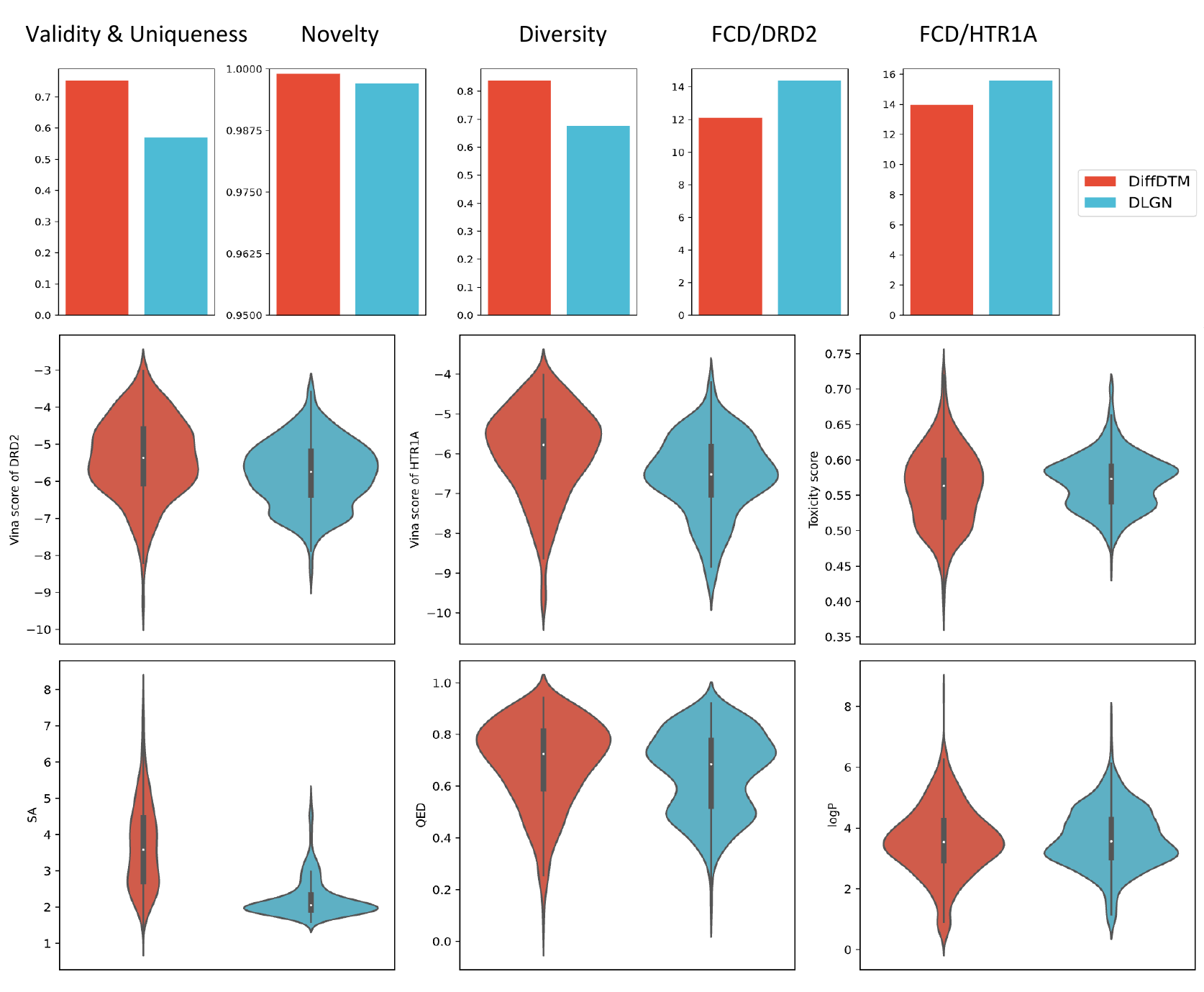}
    \caption{The results generated samples from DiffDTM and DLGN for DRD2 and HTR1A. We report the results of DiffDTM, and DLGN on validity\&uniqueness, novelty, diversity, FCD, vina score, toxicity score, sa, QED, and logP.}
    \label{fig:performance_dh}
\end{figure}

We adopt DLGN\cite{lu2021novo} which is trained on the molecules that are bioactive towards DRD2 and HTR1A for comparison. Specifically, the authors collected 344184 bioactive molecules in the ChEMBL datasets to pre-train the model to learn the basic grammar of molecule SMILES sequences and 2156 bioactive molecules for DRD2 and 2787 bioactive molecules for HTR1A to train bioactivity predictors for adversarial learning and reinforce learning. Instead, our model is only trained on our dataset and does not see DRD2 and HTR1A proteins during training. We finally generated valid 10000 molecules using each model, which is the 'cold start' strategy. We report quantitative metrics and molecule chemical properties to evaluate the generated molecules from DIffDTM and DLGN. In addition to the metrics used in general performance comparison, here we additionally report Validity\&Uniqueness, diversity, and Fréchet ChemNet Distance (FCD)\cite{preuer2018frechet}. Valid\&Uniqueness is the percentage of valid and unique molecules. Diversity represents the average pairwise Tanimoto dissimilarity of the generated molecules. FCD measures the distances between molecules and the reference set using the embeddings learned by a neural network. 

As shown in Figure \ref{fig:performance_dh}, we investigate that DiffDTM can generate more valid, unique, novel and diverse molecules compared to DLGN since DiffDTM is based on graph construction and introduce randomness via the transition probability matrices while DLGN is based on SMILES sequences and a small perturbation of a token in the sequence may cause a collapse and learned grammar is limited, leading to repeated molecules generation. In other words, DiffDTM could generate more novel molecules, indicating that DIffDTM can explore a larger chemical space to discover novel and rational structures. We also report the FCD distance between the generated molecules of the two models and the molecules in the test set of DRD2 bioactive dataset and HTR1A bioactive dataset which are split by the authors of DLGN. Interestingly, the molecules generated by DiffDTM show shorter FCD distances towards the two specific targets compared to the molecules generated by DLGN, demonstrating that the molecules generated by DiffDTM have a higher probability of being bioactive towards DRD2 and HTR1A since their chemical structures are more similar to both DRD2 bioactive dataset and HTR1A bioactive dataset.

Figure \ref{fig:performance_dh} also presents the chemical property value distributions of the generated molecules from the two models. Our observations reveal that the distribution of vina scores for DRD2 generated by DiffDTM is close to the distribution generated by DLGN. Furthermore, DiffDTM achieves a superior average vina score (-6.06) than DLGN (-5.79). Similarly, the vina scores for HTR1A of DiffDTM also show a close distribution to the vina scores of DLGN. And DiffDTM outperforms DLGN on the average vina score of HTR1A (-6.58 VS -6.50). We also observe that our model generates molecules with lower toxicity scores (0.561 VS 0.569). The molecules generated by DiffDTM exhibit higher synthetic complexity, attributable to an inclination towards complex structural features such as long chains and cyclic moieties. Notably, despite this predisposition, DiffDTM achieves superior performance in terms of the QED metric with a value of 0.691, compared to DLGN's value of 0.654. In addition, the molecules generated by DiffDTM and DLGN exhibit a reasonable distribution of logP values, indicating a proper partitioning of compounds between aqueous and lipid phases, which is usually associated with good solubility and permeability of compounds. This further suggests that the generated compounds have a high likelihood of exhibiting favourable absorption, distribution, metabolism, and excretion properties.

\subsection*{DiffDTM enables selective molecules generation towards DRD2 and HTR1A}

\begin{figure}[htbp]
    \centering
    \includegraphics{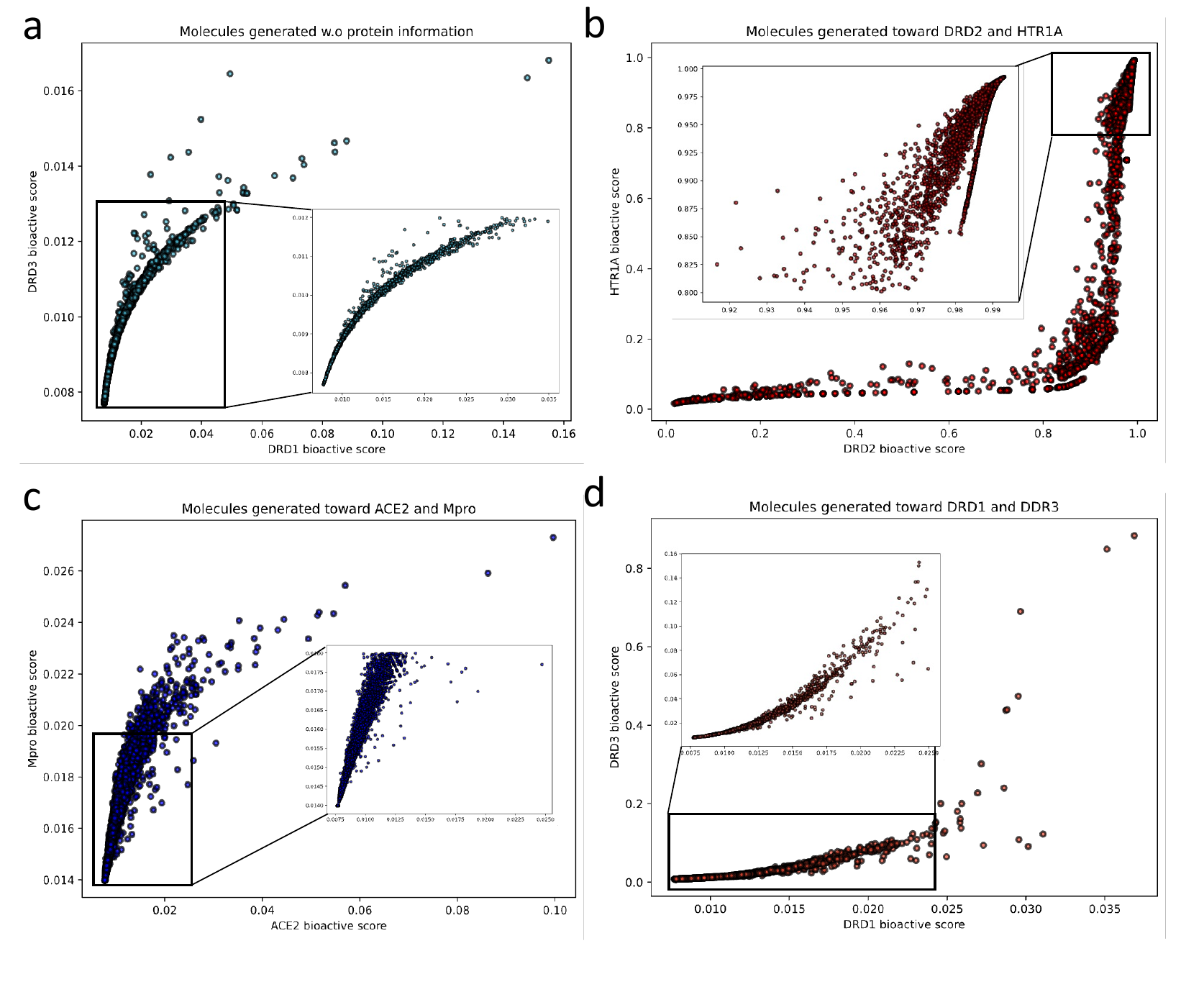}
    \caption{The bioactive probabilities of generated molecules towards specific targets that are predicted by a classification model. (a). The probabilities of Molecules generated without protein information as DRD2/HTR1A bioactive compounds. (b). The probabilities of molecules with protein information as DRD2/HTR1A bioactive compounds. (c). The probabilities of molecules with protein information as 3C like $M^\text{pro}$ and ACE2 bioactive compounds. (d). The probabilities of molecules with protein information as DRD1 and DRD3 bioactive compounds.}
    \label{fig:bioactive}
\end{figure}

In order to directly estimate whether DiffDTM can generate molecules which are bioactive towards DRD2 and HTR1A, we utilize our previous work CoaDTI\cite{huang2022coadti} which is trained on BindingDB dataset to predict whether the molecule interacts with the protein. We suppose that the compound is likely to be a dual-target drug candidate if it is classified as bioactive towards both molecules. We apply DiffDTM to generate molecules without protein information and with DRD2 and HTR1A information. Following previous work\cite{lu2021novo}, we utilize each model to generate 10000 molecules separately and plot the generated molecules' prediction probabilities outputted by CoaDTI in a plane rectangular coordinate system for intuitive visualization. On the coordinate plane, a molecule is represented by a point, with its corresponding predicted probability from the CoaDTI on the x-axis and predicted probability from the CoaDTI on the y-axis.

As shown in figure \ref{fig:bioactive}a, we observe that the generated molecules demonstrate low bioactive scores if DiffDTM does not incorporate protein information. Most of the generated molecules are predicted to have less than 0.1 probability of being bioactive towards to DRD2 and HTR1A. Although there are some outliers which have a relatively higher probability of being bioactive towards the two targets, they only account for a small fraction of the overall population. Here, we set 0.5 as the threshold to determine whether the molecules are bioactive to the target. Specifically, the mean probabilities of molecules as DRD2 and HTR1A bioactive compounds are 0.0112 and 0.0089, respectively, and there is no molecule classified bioactive towards both targets.

Figure \ref{fig:bioactive}b demonstrates the bioactive probability distribution of the generated molecules which are targeted for DRD2 and HTR1A. As we can observe, most generated molecules are predicted to have high probabilities of being bioactive to DRD2 and HTR1A. Specifically, 8769 molecules are considered to be bioactive towards both targets by CoaDTI, while only 309 molecules are classified as negative, indicating that the improvement benefited from incorporating protein information for conditional generation. 

In order to evaluate whether the generated molecules are selective towards DRD2 and HTR1A, we apply CoaDTI to predict whether the generated molecules are bioactive to another two pairs of targets. Here, we select SARS-CoV-2 3CL M$^{\text{pro}}$ and SARS-CoV-2 receptor Angiotensin-converting enzyme 2 (ACE2) as one pair of drugs, and we suppose that the generated molecules have a low probability of being positive towards the two targets. As presented in figure \ref{fig:bioactive}c, all of the molecules are not bioactive towards 3CL M$^{\text{pro}}$ and ACE2. Since we generated molecules targeting DRD2 and HTR1A, desired molecules should be selective towards the two specific targets. In general, the activity of drugs against other receptors in the dopamine system may affect the safety and efficacy of drugs. For example, if a drug has activity against both D2 and D1 receptors, it may produce competitive antagonism that unnecessarily weakens the therapeutic effect. If the drug also has activity against other dopamine receptors such as D3 and D4 receptors, it may have adverse effects on the selectivity and specificity of the drug. Therefore, we selected DRD1 and DRD3 as another pair of targets to verify whether the generated molecules exhibit low activity against other dopamine receptors. In Figure \ref{fig:bioactive}d, it can be observed that the majority of the generated molecules are classified as non-bioactive towards DRD1 and DRD3, with only three molecules being bioactive towards DRD3. Overall, the results suggest that DiffDTM has the ability to generate bioactive and selective molecules for desired protein targets.

\subsection*{DiffDTM can generate molecules which are similar to FDA-approved drugs with corroboration of binding affinities}
DiffDTM is designed and envisioned to generate bioactive drug candidates for specific targets to modulate biology functions. Therefore, we calculate the structural similarities between generated molecules and FDA-approved drugs to confirm DiffDTM's capacity of generating rational molecule candidates. Following previous work\cite{lu2021novo}, we collect 2621 FDA-approved small molecule drugs from DrugBank\cite{wishart2018drugbank}. We also utilize 10000 generated molecules targeted for DRD2 and HTR1A to verify the drug generation potential of DiffDTM. We calculate the Tanimoto similarities between every generated molecule and the drugs approved by FDA and finally pick out the top 50 pairs to study the pharmacological mechanism.  

\begin{figure}[htbp]
    \centering
    \includegraphics{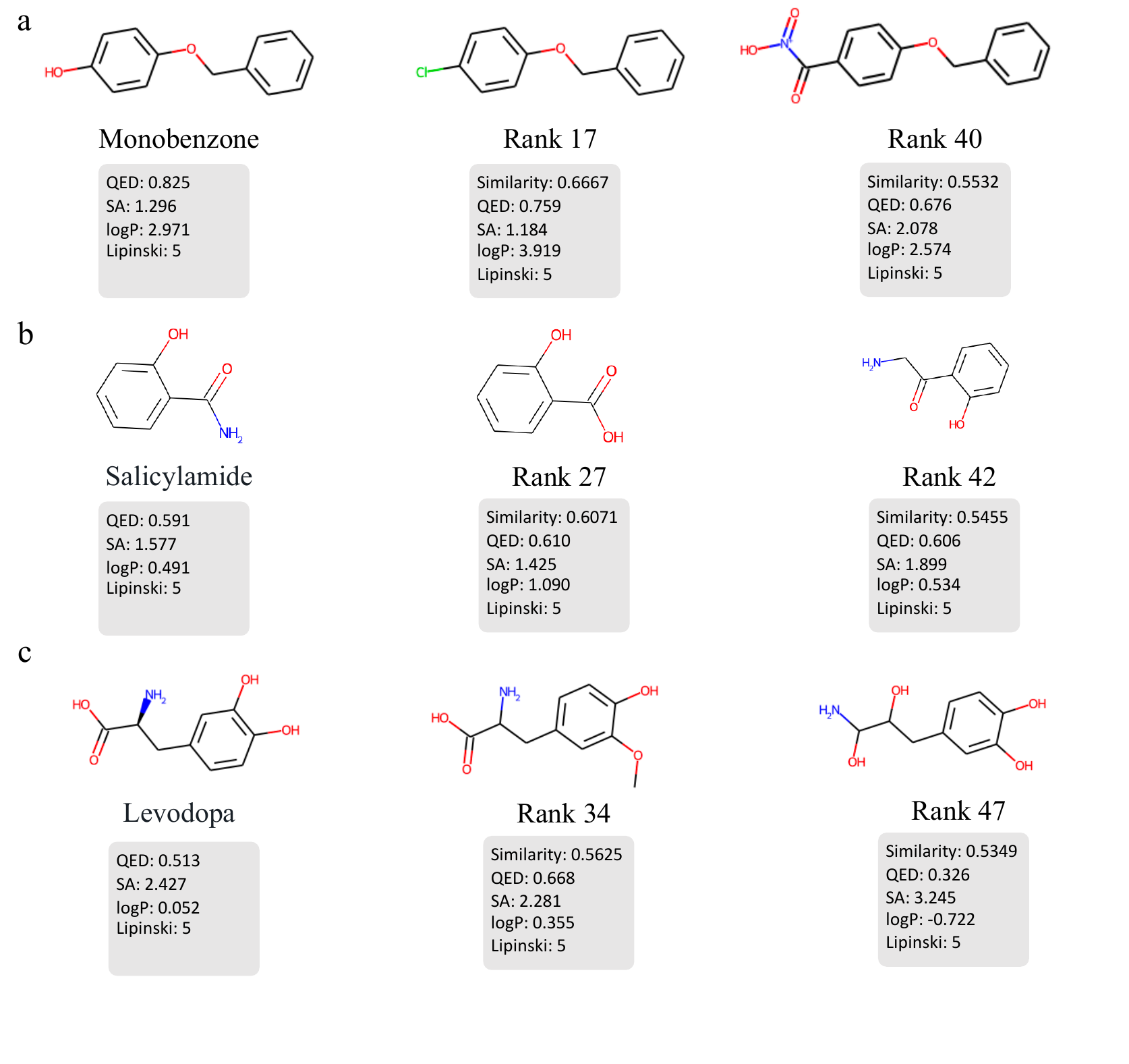}
    \caption{The comparison of FDA-approved molecules (left) and two molecules (middle, right) generated by DiffDTM among the top 50 similarity pairs. (a). Monbenzone. (b). Salicylamide. (c). Levodopa.}
    \label{fig:similarity}
\end{figure}

It is notable that our model "cold starts" for DRD2 and HTR1A, which means that our model does not see the FDA-approved drugs in the training set. On the contrary, DLGN fine-tunes their model on the bioactive molecule data towards both targets. Table \ref{Tab: similarity} reports the statistical results of the top 50 similar generated molecules from DiffDTM and DLGN. It can be observed that DiffDTM achieves comparable performance compared to DLGN and can even generate ten molecules identical to the FDA-approved drug. We list the example generated molecules which are similar to FDA-approved drug molecules (figure \ref{fig:similarity}). For example, DiffDTM generates two molecules which are ranked 17 and 40 (figure \ref{fig:similarity}a) with similarities of 0.6667 and 0.5532 to the monobenzyl ether of hydroquinone used medically for depigmentation termed Monobenzone. Among them, the molecule ranked 17 is easier to synthesize, and the SA score of Monobenzone is 1.296 while the SA score of this molecule is 1.184. There are also two molecules (rank 27 and 42 in figure \ref{fig:similarity}b) with similarities of 0.6071 and 0.5455 to the non-prescription drug with analgesic and antipyretic properties termed Salicylamide. The molecule ranked 27 has a higher QED value and lower SA score than Salicylamide. 

\begin{figure}[htbp]
    \centering
    \includegraphics{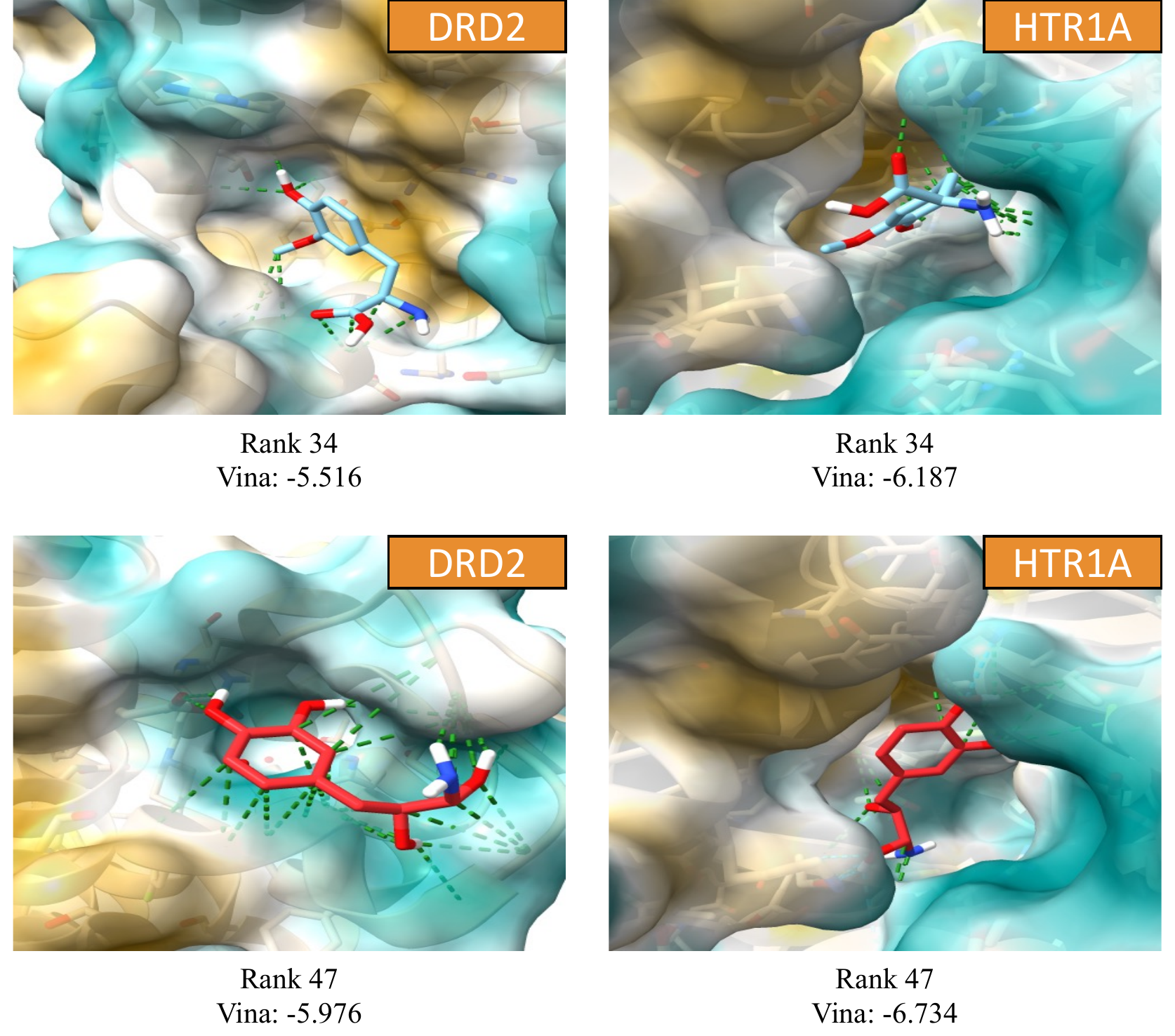}
    \caption{3D visualization of ligands (ranked 34 and 47) docked against DRD2 and HTR1A}
    \label{fig:similarity_vina}
\end{figure}

Interestingly, DiffDTM generates molecules (rank 34 and rank 47 in figure \ref{fig:similarity}c) that are similar to a dopamine precursor used in the management of Parkinson's disease termed Levodopa, which targets DRD2. Similarly, the molecule ranked 34 has a higher QED value and lower SA score than Levodopa. We further utilize RDKit to predict the conformations of molecules ranked 34 and 47, and then employ Gnina to obtain the docking poses towards DRD2 and HTR1A with vina scores. The 3D binding poses produced by Gnina for the two molecules for DRD2 and HTR1A demonstrate that they fit well into the protein binding pocket and promote favourable ligand-protein interactions (figure \ref{fig:similarity_vina}).

\begin{table}[htb]
\centering
\caption{Statistical information of the top 50 similar pairs. \label{Tab: similarity}}
\scalebox{1}{
\begin{tabular}{lccc}\\
\toprule 
Methods & Min similarity & Average similarity & Number of identical molecules\\
\midrule        
DiffDTM & 0.5333 & 0.6925 & 10\\
DLGN & 0.6333 & 0.7428 & 8\\
\bottomrule
\end{tabular}}
\end{table}

\section*{Conclusion}
\textit{De novo} drug molecule generation targeted for multiple proteins is increasingly important due to the difficulty of healing and alleviating complex disorders resulting from multiple genetic and/or environmental factors that lead to the breakdown of robust physiological systems. The deep generative models which are based on deep learning technologies have paved the way for efficient and accurate molecule generation. However, the existing methods are always trained on the bioactive data towards specific targets and cannot generalize to new dual targets. In this paper, we proposed DiffDTM, a novel conditional structure-free deep generative model based on Diffusion model for Dual Targets based Molecule generation to tackle the above issues and improve the generation qualities. In particular, DiffDTM learns an information fusion function to enable the molecule representation to incorporate protein information to implement conditional generation. By incorporating the pre-trained model ESM2 and graph transformer, DiffDTM elucidates a suitable path towards learning the feature representations of drug modalities and protein modalities in the latent space. Based on the diffusion model, DiffDTM can gradually remove the noise in each time step to obtain a bioactive molecule targeted for inputted proteins. The extensive experiments demonstrate that it achieves substantially better or competitive performance against the SOTA methods. Furthermore, we utilize DiffDTM to generate bioactive molecules towards DRD2 and HTR1A to verify its potential on dual-target drug candidate design. The statistical and chemical property results indicate that DiffDTM can be easily generalized into unseen dual targets to generate bioactive molecules with high affinities to DRD2 and HTR1A which are similar to the approved molecule drugs, addressing the issues of requiring insufficient active molecule data for training and the need to retrain when encountering new targets. We speculate that the use of DiffDTM to generate bioactive molecules towards desired targets may lead to the acceleration of de novo drug discovery for different diseases. 

\section*{Method}
\subsection*{Data sources}
The pair of molecules and targets index were downloaded from the BindingDB database (https://www.bindingdb.org) and further extracted information of the specified human kinases\cite{uniprot2021uniprot}. The experimental bioactivity data (IC50, Kd, Ki, EC50) were queried from this database, and the same character information were further collected from six databases, including the dataset of Davis\cite{davis2011comprehensive}, SARfari\cite{davies2015adme}, Metz\cite{metz2011navigating}, PKS1\cite{elkins2016comprehensive}, PKS2\cite{drewry2017progress}. The FASTA information of protein was collected from UniProt website (https://www.uniprot.org). The above data set is processed as follows: (1). The compounds with corresponding biological activity data or canonical SMILES string were kept, furthermore, the ions and additional salts and solvents in the molecule system were deleted using RDKIT pacakage\cite{landrum2013rdkit}. (2). The compound-kinase pairs with both positive and negative labels in all datasets were eliminated due to data conflict. The drug molecule is bioactive to the protein if IC50 is less than 100nm or negative if IC50 is greater than 10,000nm. (3). After the deletion of mutant kinases and kinases without both active and inactive data points, we obtained over 139,330 bioactivity molecule-protein pairs. Then we filtered the molecules which are bioactive towards at least two proteins. Most molecules are bioactive towards two protein targets while we combine the proteins whose e-value from BLAST\cite{camacho2009blast+} search results is larger than 0.4 of the molecules which bioactive towards more than two protein targets to avoid data overwhelming. The filtered dataset contains 160,266 molecule-dual-targets pairs, which have 29,463 molecules and 1636 proteins. For the test set, we utilized 100 molecule-dual targets which the proteins were not included in the training set and BLAST e-values of the proteins in the test set and training set were less than 0.4 to avoid data leakage.

\subsection*{Metrics}
We adopt widely-used three quantitative metrics and six chemical property metrics\cite{polykovskiy2020molecular}\cite{luo20213d}\cite{huang2022mdm} to evaluate the quality of molecules generated by DiffDTM: (1)\textbf{Validity} is the percentage of valid molecules  that follow
the chemical valency rules specified by RDkit; (2)\textbf{Uniqueness} is the percentage of unique molecules in valid molecules; (3)\textbf{Novelty} is the percentage of generated valid molecules that do not exist in the training set; (4)\textbf{Vina Score} estimates the binding affinity between the ligand and the target pocket which is the most important measurement to evaluate how the generated molecule fits into the protein pocket of interest; (5)\textbf{Toxicic socre} is the value outputted by eToxPred\cite{pu2019etoxpred} that indicate the toxicity probability; (6)\textbf{SA} (synthetic accessibility) measures the molecule synthetic accessibility; (7)\textbf{QED} estimates the drug-likeness of the molecule via combining several desirable molecular properties; (8)\textbf{LogP} indicates the octanol-water partition coefficient, which should be between -0.4 and 5.6 if the molecule is a good drug candidate \cite{ghose1999knowledge}; (9)\textbf{Lipinski}    
measures how many rules the drug follows five Lipinski’s rules \cite{lipinski2012experimental}. In order to extensively compare the quality of generated molecules towards DRD2 and HTR1A, we additionally report three metrics: (1)\textbf{Validity\&Uniqueness} is the percentage of valid and unique molecules; (2) \textbf{Diversity} represents the average pairwise Tanimoto dissimilarity of the generated molecules ; (3)\textbf{Fréchet ChemNet Distance (FCD)}\cite{preuer2018frechet} measures the distances between molecules and the reference set using the embeddings learned by a neural network. 
\subsection*{Baseline models}
For general metrics evaluation, we selected GraphVAE, JTVAE, and MolGPT as baseline models. For generating bioactive molecules towards DRD2 and HTR1A, we selected DLGN as the baseline model for comparison.
\begin{itemize}
\item \textbf{GraphVAE}: We utilized a conditional GraphVAE to learn the molecular graphs distribution and generate the molecular graphs given conditional protein information. 
\item \textbf{JTVAE}: We revised JTVAE that first generates a scaffold junction tree and then assembles nodes in the tree into a molecular graph to make it accepts protein information as inputs as a conditional generative model.
\item \textbf{MolGPT}: We revised MolGPT that generates Molecular SMILES sequences using a transformer-decoder model to make it accepts protein information as inputs as a conditional generative model.
\item \textbf{DLGN}: A generative model that combines adversary learning and reinforcement learning to generate molecule sequences towards to DRD2 and HTR1A.

\end{itemize}

\subsection*{Preliminary}
Let $G = (X,E)$ denote the molecular graph where $X = (x_1,x_2,\cdots,x_n)\in \{0,1\}^{n \times f}$ denotes the discrete one-hot encoded atom types (a.k.a, chemical elements) and atom features, and $E \in \mathbb{R}^{n \times n \times b}$ represents the undirected edges between atoms which are chemical bonds grouped the one-hot encoding $e_{ij}$ for each edge. $A$ denotes the adjacent matrix. Besides, we denote protein amino acid sequence as $\mathcal{P}=\{a_{1},a_{2},\dots,a_{n}\}$ where $a_{i}$ is the $i^{th}$ index of the 23 amino acids. We denote $\mathcal{G}_t$ for $t=1,\dots, T$ as a sequence of latent geometries where $t$ indicates the index of diffusion steps.

\subsection*{Discrete diffusion model}
The diffusion model \cite{sohl2015deep} can be formulated as two Markov chains: {\em diffusion process} and {\em reverse process}. The diffusion process iteratively adds noise to corrupt the input data point $G^0$ to a predefined noise distribution $G^1,\dots,G^T$ while the reverse process applies a noise model to predict the noise and then gradually removes the noise to recover the real data point from noisy data. 

In this work, we closely follow the developed discrete diffusion model\cite{vignac2023digress}\cite{austin2021structured} since atom types and chemical bonds are all discrete variables in the model. The discrete diffusion model defines a discrete transition matrix $\overline{Q}^t = Q^1Q^2 \dots Q^t$. Then we could define the posterior distribution given protein information $q(G^{t-1}|G^t,G^0,\mathcal{P})$ as a closed form:
\begin{equation}
    q(G^{t-1}|G^t,G^0,\mathcal{P}) \propto G^t(Q^t)^\prime \odot G^0 \overline{Q}^{t-1}
\end{equation}

where $\odot$ denotes the pairwise product and $(Q^t)^\prime$ is the transpose of $Q^t$. For discrete nodes and edges, the transition probabilities are defined by two matrices: $[Q^t_X]_{ij} = q(X^t=j|X^{t-1}=i)$ and $[Q^t_E]_{ij} = q(E^t=j|E^{t-1}=i)$. Then the diffusion process could be defined as:
\begin{equation}
    q\left(G^{t} \mid G^{t-1},\mathcal{P},\right)=\left({X}^{t-1} {Q}_{X}^{t}, \mathbf{E}^{t-1} {Q}_{E}^{t}\right) \text { and } q\left(G^{t} \mid G^0,\mathcal{P},\right)=\left({X} \overline{{Q}}_{X}^{t}, \mathbf{E} \overline{{Q}}_{E}^{t}\right)
\end{equation}

Different from the reverse process in continuous diffusion models utilizing a neural network $\phi_\theta$ to predict the discrete noise $\epsilon_\theta$, the discrete diffusion model utilizes $\phi_\theta(G^t,\mathcal{P})$ to directly predict the less noisy molecule graphs $G^{t-1}$ from noisy molecule graphs $G^{t}$. To train this neural network, we adopt cross-entropy loss between the predicted probabilities $\hat{p}^{G}=\left(\hat{p}^{X}, \hat{p}^{E}\right)$ and true molecular graph $G$ given protein information as the objective function to optimize:
\begin{equation}
    l\left(\hat{p}^{G}, G\right)=\sum_{1 \leq i \leq n} \operatorname{cross-entropy}\left(x_{i}, \hat{p}_{i}^{X}\right)+\lambda \sum_{1 \leq i, j \leq n} \operatorname{cross-entropy}\left(e_{i j}, \hat{p}_{i j}^{E}\right)
\end{equation}

Then the reverse process could utilize the trained network to obtain the transition probability $p_\theta(G^{t-1}|G^t)$, which is producted over nodes and edges:
\begin{equation}
    p_{\theta}\left(G^{t-1} \mid G^{t}, \mathcal{P}\right)=\prod_{1 \leq i \leq n} p_{\theta}\left(x_{i}^{t-1} \mid G^{t}, \mathcal{P}\right) \prod_{1 \leq i, j \leq n} p_{\theta}\left(e_{i j}^{t-1} \mid G^{t}, \mathcal{P}\right)
\end{equation}

Then this formulation could be marginalized over the network predictions:
\begin{equation}
    p_{\theta}\left(x_{i}^{t-1} \mid G^{t}\right)=\int_{x_{i}} p_{\theta}\left(x_{i}^{t-1} \mid x_{i}, G^{t}\right) d p_{\theta}\left(x_{i} \mid G^{t}\right)=\sum_{x \in \mathcal{X}} p_{\theta}\left(x_{i}^{t-1} \mid x_{i}=x, G^{t}\right) \hat{p}_{i}^{X}(x)
\end{equation}
where
\begin{equation}
    p_{\theta}\left(x_{i}^{t-1} \mid x_{i}=x, G^{t}\right) = 
    \left\{\begin{array}{ll}
    q(x^{t-1}_{i}|x_{i} = x, x^{t}_{i}), 
    & \text{if }q(x^{t-1}_{i}|x_{i} = x)>0 \\
    0, &\text{otherwise}
    \end{array}\right.
\end{equation}
Similarly, $p_\theta(e^{t-1}_{ij}|e^t_{ij})$ has the same formulation.

\subsection*{ESM2 for protein encoding}
In order to obtain protein enriched representation, we utilize a pre-trained model ESM2 to encode protein sequences. Proteins are represented as amino acid sequences which are analogous to the natural text with limited sequences. First, we tokenize the protein sequences by the predefined tokenizer of ESM2. Then we input the tokenized protein sequences into the ESM encoder. The attention layer in ESM could capture the short and long distance dependency of the residues, which can reflect the 3D structure to a certain extent. We utilize the learned representation of the last layer as the protein information: $[p^{cls},p^s,p^{sep}] = \text(ESM2)(\mathcal{P})$, where $p^{cls}$ entails the semantic information of the entire protein sequence, $p^s$ is the pooled protein sequence vectors and $p^{cls}$ is the separator vector. Finally, we feed $p^{cls}$ and $p^s$ into the subsequent modules according to the information fusion strategies.

\subsection*{Information fusion modules}
We have designed three information fusion strategies cross-attention, feature concatenation and virtual nodes to fuse the protein information and molecule information. The cross-attention consists of two transformer layers. The first transformer layer encodes the atom features of the molecule which we treat the atom features as the sequence data. The attention mechanism can learn the effects of the nodes in different positions. The encoding process of transformer is formulated as follows:
\begin{equation}
    x_{\text{mol}}=\text{LayerNorm}(x_{\text{emb}}+\text{FFL}(\text{MHA}(x_{\text{emb}}))),
\end{equation}
where FFL is the feed-forward layer, MHA is the multi-head self-attention layer. Given the dense features of molecules $x_{\text{mol}}$ and proteins ${p}^{s}$, the next step is to utilize the second transformer to join the two modalities. The input of scaled dot-product attention consists of three matrixes, queries, keys and values, which is formulated as followings:
\begin{equation}
        Q,K,V = \text{Linear}_Q\left(X_{\text{emb}}\right),\text{Linear}_K\left(X_{\text{emb}}\right),\text{Linear}_V\left(X_{\text{emb}}\right), 
\end{equation}
where $X_{\text{emb}} = \mathbf{X}_d$ when feeding drug embeddings and $X_{\text{embedding}} = \mathbf{X}_p$ when feeding protein embeddings.
Based on the three matrices, the attention score can be calculated by:
\begin{equation}
    \text{Attention}\left(Q,K,V\right)\ =\ softmax\left(\frac{QK^T}{\sqrt{d}}\right)V,
\end{equation}
where $\sqrt{d}$ turns the attention matrix into a standard normal distribution.

To learn the information from different feature subspaces at different positions, multi-head attention is incorporated into the deep Co-attention module. Multi-head attention consists of $h$ heads and outputs the values. The values are then concatenated and projected again. 
\begin{equation}
    \begin{split}
        \text{MHA}(Q,K,V) = \text{CONCAT}(\text{head}_1, \text{head}_2, \dots, \text{head}_h)W^O, \\
        \textbf{where} \ \text{head}_i = \text{Attention}(QW_i^Q,KW_i^K,VW_i^V), 
    \end{split}
\end{equation}
where $W_i^Q, W_i^K, W_i^V \in \mathcal{R}^{d \times d_h}$ are the projection matrics for the $i^{th}$ head and $W^O \in \mathcal{R}^{hd_h \times d}$. Here, $d_h$ is the dimension of the output from each head, and we employ $d_h = d/h$. To learn the intermodality information, we input the drug as query $Q$, and protein data as key $K$ and value $V$. This can measure the effects of proteins on the drug. Finally, we utilize the outputted molecule representation containing protein information $x^\prime_\text{mol}$ to the subsequent graph transformer layer.

For the concatenation strategy, we follow previous work to parameterise the conditioning by concatenating the drug embeddings $x_\text{emb}$ and protein embeddings $p^s$ in the feature dimension: $x^\prime_\text{mol} = [x_\text{emb},p^s, \text{dim='feature'}]$. For the virtual node strategy, we treat the protein embeddings as a virtual node. Then we concatenate the protein embeddings and drug embeddings in the node dimension: $x^\prime_\text{mol} = [x_\text{emb},p^s, \text{dim='node'}]$. In order to learn the neighbour information via the message passing mechanism in the subsequent graph neural network, we construct virtual edges to connect the virtual node and atom nodes.

\subsection*{Graph prediction neural network}
In order to predict the real molecule graph, we employ graph transformer\cite{dwivedi2020generalization} as the prediction neural network. Each graph transformer layer consists of a graph attention module and a fully-connected layers following layer normalization. The updation of each graph transformer layer is formulated as follows:

\begin{equation}
\hat{h}_{i}^{\ell+1}=O_{h}^{\ell} \|_{k=1}^{H}\left(\sum_{j \in \mathcal{N}_{i}} w_{i j}^{k, \ell} V^{k, \ell} h_{j}^{\ell}\right)
\text{where},  w_{i j}^{k, \ell}=\operatorname{softmax}_{j}\left(\frac{Q^{k, \ell} h_{i}^{\ell} \cdot K^{k, \ell} h_{j}^{\ell}}{\sqrt{d_{k}}}\right)
\end{equation}
$Q^{k, \ell}, K^{k, \ell}, V^{k, \ell} \in \mathbb{R}^{d_{k} \times d}, O_{h}^{\ell} \in \mathbb{R}^{d \times d}, k=1 \text { to } H$ denotes the number of attention  heads, and $\|$ denotes concatenation process. Then the attention outputs $\hat{h}_{i}^{\ell+1}$ are fed into a Feed Forward layer with the residual connections and layernorm layers \cite{ba2016layer}.
\begin{equation}
    h_{i}^{\ell+1}=\text{LayerNorm}(x_{\text{emb}}+\text{FFL}(\hat{h}_{i}^{\ell+1}))
\end{equation}

The enriched molecule atom representation $x^\prime_\text{mol}$ is inputted to the first layer of the graph transformer as $h^0$. Finally, we could obtain the atom feature vectors $h^\prime$

Regarding the edge features, they can be considered as pairwise scores corresponding to a node pair. Thus, it requires pairwise attention to calculate the edged features to implicit edge scores. Rather than use $w_{i j}^{k, \ell}$ for node features, the edge feature updation layer utilizes the intermediate weight score $\hat{w}_{i j}^{k, \ell}$ and inject the available edge information by multiplying $\hat{w}_{i j}^{k, \ell}$ and $e_{ij}$. It is natural to utilize the edge information in a graph rather than in Natural Language process since there is little information between two words. The update of each graph transformer layer for edge features and the new weight score that contains the edge information is formulated as follows:
\begin{equation}
    \begin{aligned}
\hat{e}_{i j}^{\ell+1} & =O_{e}^{\ell} \|_{k=1}^{H}\left(\hat{w}_{i j}^{k, \ell}\right), \text { where } \\
w_{i j}^{k, \ell} & =\operatorname{softmax}_{j}\left(\hat{w}_{i j}^{k, \ell}\right) \\
\hat{w}_{i j}^{k, \ell} & =\left(\frac{Q^{k, \ell} h_{i}^{\ell} \cdot K^{k, \ell} h_{j}^{\ell}}{\sqrt{d_{k}}}\right) \cdot E^{k, \ell} e_{i j}^{\ell}
\end{aligned}
\end{equation}
Finally, we could obtain the edge feature vectors $e^\prime$ from the last graph transformer layer. We employ two MLPs to predict the atom types and chemical bonds of the denoised molecular graph:
\begin{equation}
    X = \text{MLP}_x(h^\prime), E = \text{MLP}_e(e^\prime)
\end{equation}

\subsection*{Sampling process}
Since we have formulated the model of $p_\theta$, now we can calculate the denoised atom types $\hat{p}^x$ and chemical bonds $\hat{p}^E$. As presented in Figure \ref{fig:model}a, the chaotic state $G_T$ is sampled from a predefined distribution $q_x(n) \times q_e(n)$. The next less chaotic state $G_{T-1}$ is generated by $\prod_{1 \leq i \leq n} p_{\theta}\left(x_{i}^{t-1} \mid G^{t}, \mathcal{P}\right) \prod_{1 \leq i, j \leq n} p_{\theta}\left(e_{i j}^{t-1} \mid G^{t}, \mathcal{P}\right)$ given protein information. The final molecule $G_0$ is generated by progressively sample $G_{t-1}$ for $T$ times. Finally, the atom types and chemical bonds of the molecule are identified by adopting the argmax function to choose the atom type which has the largest value.  

\bibliography{sample}








\end{document}